\documentclass[conference]{IEEEtran}
\IEEEoverridecommandlockouts
\usepackage{cite}
\usepackage{amsmath,amssymb,amsfonts}
\usepackage{algorithmic}
\usepackage{graphicx}
\usepackage{textcomp}
\usepackage{xcolor}
\usepackage{subcaption}
\def\BibTeX{{\rm B\kern-.05em{\sc i\kern-.025em b}\kern-.08em
    T\kern-.1667em\lower.7ex\hbox{E}\kern-.125emX}}
\begin{document}
\title{Enhancing Otological Surgery: Co-Designing a Parallel Robot with Surgeon Input}

\author{Durgesh Haribhau Salunkhe$^{1}$, Guillaume Michel$^{1, 2}$, Shivesh Kumar$^{3, 4}$ and Damien Chablat$^{1}$
\thanks{$^{1}$Nantes Université, École Centrale Nantes, CNRS, LS2N, UMR 6004, 44000 Nantes, France {\tt\small \{durgesh.salunkhe, damien.chablat, guillaume.michel\}@ls2n.fr}}
\thanks{$^{2}$Centre Hospitalier Universitaire de Nantes, Nantes, France, 44000}
\thanks{$^3$Dynamics Division, Department of Mechanics and Maritime Sciences, Chalmers University of Technology, Gothenburg, Sweden {\tt\small shivesh@chalmers.se}}
\thanks{$^4$Robotics Innovation Center, German Research Center for Artificial Intelligence, Bremen, Germany}%
}

\maketitle
\graphicspath{{./images/}}

\begin{abstract}
This work presents the development of a parallel manipulator used for otological surgery from the perspective of co-design. Co-design refers to the simultaneous involvement of the end-users (surgeons), stakeholders (designers, ergonomic experts, manufacturers), and experts from the fields of optimization and mechanisms. The role of each member is discussed in detail and the interactions between the stakeholders are presented. Co-design facilitates a reduction in the parameter space considered during mechanism optimization, leading to a more efficient design process. 
Additionally, the co-design principles help avoid unforeseen errors and help in quicker adaptation of the proposed solution.
\end{abstract}

\section{Introduction}
\begin{figure}[htbp]
    \centering
    	\begin{subfigure}{0.48\columnwidth}
    		\centering
    		\includegraphics[width=1.5 in,height = 1.35 in]{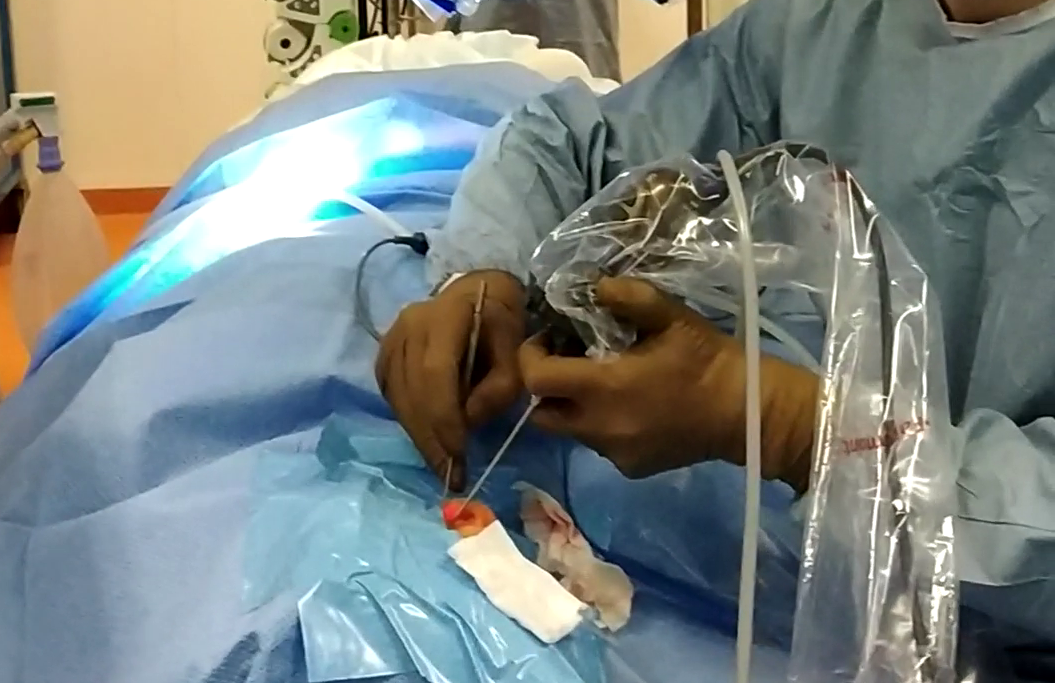}
    		\caption{The use of endoscope limits the number of instruments}
    		\label{fig:endo_1hand}
    	\end{subfigure}
    	~
    	\begin{subfigure}{0.48\columnwidth}
    		\centering
    		\includegraphics[width=1.5 in,height = 1.35in,angle =0]{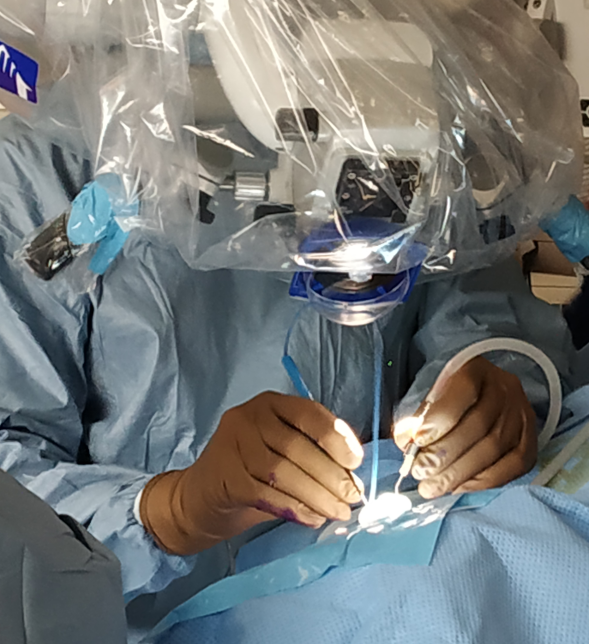}
    		\caption{Surgeon using 2 instruments in microscopic surgery}
    		\label{fig:micro_2hand}
    	\end{subfigure}
    	\caption{The comparison of the number of instruments used simultaneously while using an endoscope and a microscope.}
    	\label{fig:endo_micro_hand}
    \end{figure}

The utilization of an endoscope during otological surgery yields significant advantages in terms of visibility and access to the surgical site (see Figure \ref{fig:endo_micro_hand}). However, its constant manual manipulation by the surgeon, as depicted in Figure \ref{fig:endo_1hand}, introduces significant challenges. This constant handling of the endoscope, coupled with the necessity to switch between instruments for operating and managing bleeding in the ear, renders endoscopic surgeries laborious. Implementing a robotic arm to manipulate the endoscope as required can substantially enhance the efficacy of otological surgeries. Integration of assistive systems has the potential to drastically reduce operating time and yield positive outcomes for both the surgeon and the patient.

\begin{figure}[htbp]
    \centering
    \includegraphics[width=0.6\columnwidth]{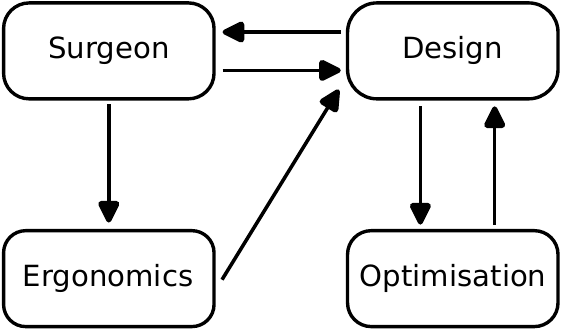}
    \caption{Interactions and knowledge transfer between experts.}
    \label{fig:stakeholders}
\end{figure}

In this work, we proposed a solution for a robot assistant by keeping all the stakeholders in the loop. The surgeons, designers, ergonomic experts, and optimization experts communicated to discuss the requirements, possible solutions, and possible improvements. Figure \ref{fig:stakeholders} shows the interactions between experts in different fields. The arrows suggest the general flow of knowledge transfer between two experts. 
\subsection{Surgeon and design expert}
\begin{figure}[htbp]
    \centering
    \includegraphics[width=\columnwidth]{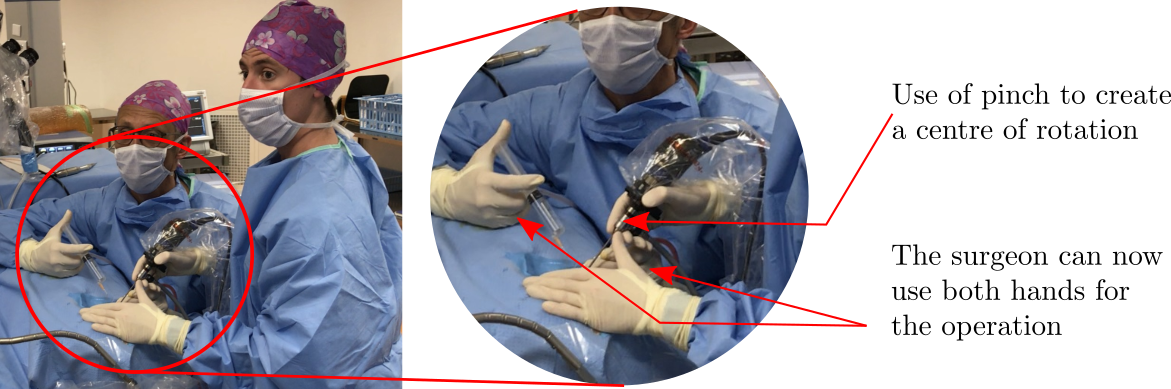}
    \caption{The surgeon demonstrates the expected output from the robot assistant to the design expert.}
    \label{fig:robot_assistant}
\end{figure}

The surgeon demonstrated the issues to the designers and proposed a rough idea of the expected output. Figure \ref{fig:robot_assistant} shows the illustration of the concept.
The designer then converted the surgeon's needs into technical specifications, such as the center of rotation of the endoscope and the types of degrees of freedom required. Extensive research on the workspace for otological surgery was conducted \cite{michel_atlas, michel_lit_review} in collaboration, leading to a better understanding of the exact requirements that must be addressed while proposing a mechanism. 
\subsection{Surgeon and ergonomic expert}
\begin{figure}[htbp]
    \centering
    \includegraphics[width=0.8\columnwidth]{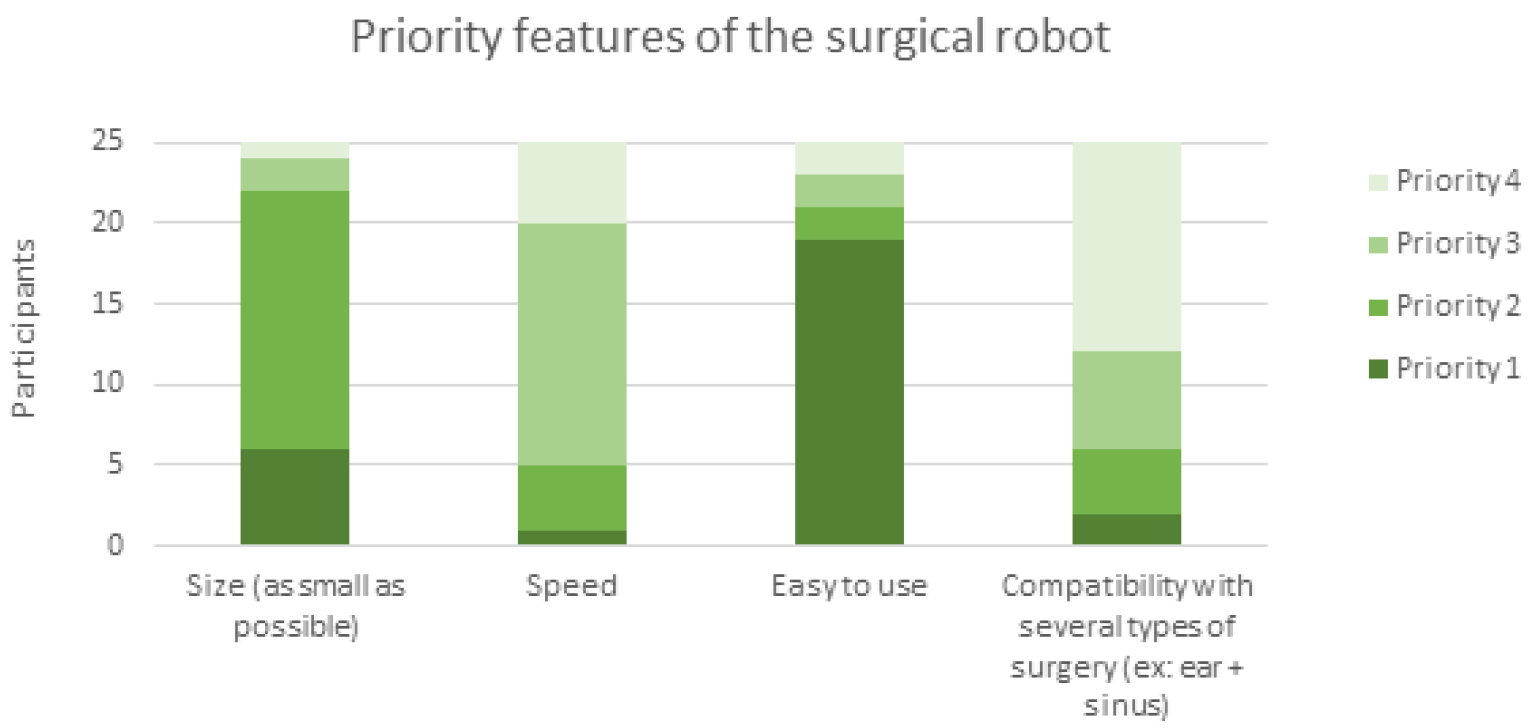}
    \caption{The priorities set by surgeons for different requirements from a robot assistant.}
    \label{fig:questionnaire_results}
\end{figure}

The interactions between the surgeon and the ergonomics expert were unidirectional, as the surgeon only gave the feedback necessary to evaluate the ergonomic measures. In this case, the ergonomic expert created a questionnaire to gain insight into the surgeon's desired speed and priority for a potential surgical robot and the device's perceived usability. Within the frame of an iterative and incremental development, this information provided insights for the design conception before building a prototype. Figure \ref{fig:questionnaire_results} shows the result from the two sets of questionnaires.
\subsection{Ergonomic expert and design expert}
\begin{figure}[htbp]
    \centering
    \includegraphics[width = 0.5\columnwidth]{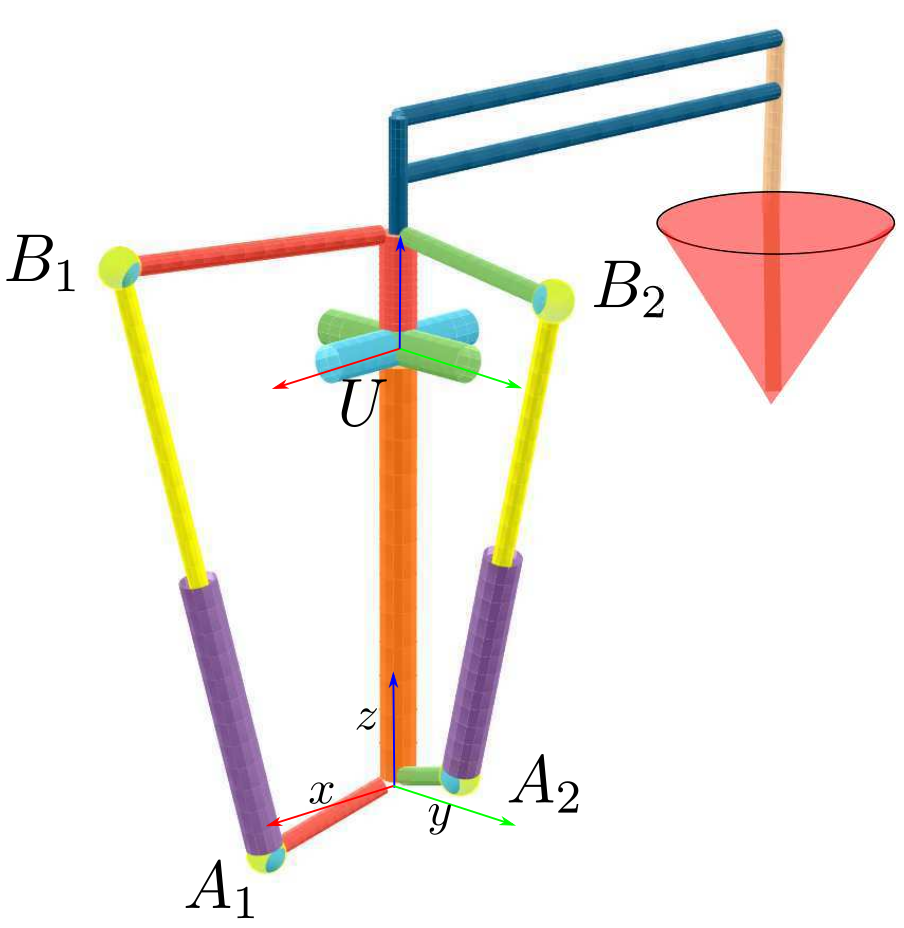}
    \caption{Schematic of the proposed 2U\underline{P}S + 1U mechanism}
    \label{fig:2UPSpara}
\end{figure}

The design expert then studied the ergonomic expert's results. The questionnaire's priorities reflected different design choices to be made while proposing the type of mechanism. For example, as the ease of use and size of the mechanism were given higher priorities, a parallel mechanism with the remote center of motion was proposed. Later, the choice of prismatic joints over revolute joints was made as the resulting mechanism reduced the occupied volume. The proposed 2 U\underline{P}S + 1U mechanism used a motion constraint generator (a universal joint) with 2 rotational degrees of freedom, and prismatic joints for actuation. The schematic of the proposed type of mechanism is shown in Figure \ref{fig:2UPSpara}.
\subsection{Design expert and optimization expert}
\begin{figure}[h!]
    \centering
    \includegraphics[width=0.6\columnwidth]{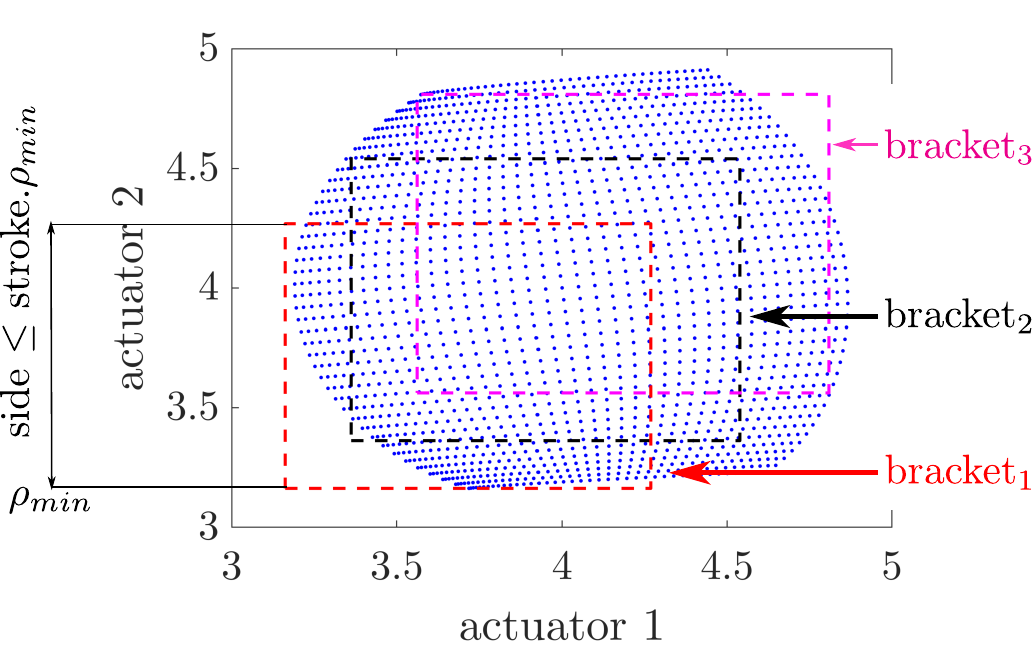}
    \caption{Different search brackets within the actuator space. The dots correspond to the pair of lengths of actuators for a configuration in a reduced dextrous workspace \cite{salunkhe_mmt22a}.}    	
    \label{fig:actuator_search}
\end{figure}

The interactions between the design experts and optimization experts were bidirectional. This is due to the overlap in expertise and the strong interdependence between chosen design parameters, objective function, and the final output. Different objective functions were studied to understand the effect of length and the placement of joints. The design optimization was done by implementing the Nelder-Mead, derivative-free optimization algorithm. The initial dimension of the optimization space was 13 as there were 4 joints to be placed ($x, y, z$ coordinates), and the height of the universal joint was used as a motion constraint generator. After the first optimization stage, the design expert noted that the optimized design parameters could be reduced to four by further constraining the placement of joints. With this insight, the optimization space was reduced to four dimensions. This drastically improved the optimization results and allowed more complex constraints to be incorporated.
Another important advantage of these interactions was the incorporation of prismatic joint limits in the optimization methodology. Figure \ref{fig:actuator_search} shows an example of how the size of the actuators influences the feasible workspace. More details on this can be found in \cite{salunkhe_mmt22a}.
\\~\\
It was further noticed that the ergonomic expert's results had to be converted into design constraints before being passed on to the optimization expert. The interactions between the surgeon and the optimization expert were similar. The design expert interacted with the surgeon, the ergonomic expert, and the optimization expert to conceptualize the issue, design constraints, and improve the optimization algorithm. The singularity analysis of the mechanism was studied in \cite{michel_jha_swami}, and the results from the interaction of the surgeon and the design expert are presented in \cite{michel_surgical_innovation}.

\bibliographystyle{unsrt}
\bibliography{references}
\end{document}